%% file: gradcam.tex
\documentclass[final]{gradcam}
\usepackage{times}
\usepackage{epsfig}
\usepackage{graphicx}
\usepackage{amsmath}
\usepackage{amssymb}
\usepackage[numbers]{natbib}
\usepackage{import}

\usepackage{multirow}
\usepackage{threeparttable}
\usepackage{algorithm, algorithmic}

\usepackage[pagebackref=true,breaklinks=true,colorlinks,bookmarks=false]{hyperref}

\renewcommand{\paragraph}[1]{{\vspace{+0mm}\noindent\bf #1}.}

\begin{document}


\title{How to Design a Compact High-Throughput Video Camera?}

\vspace{-2mm}

\author{Chenxi Qiu \qquad Tao Yue \qquad Xuemei Hu \\
Nanjing University\\
{\tt\small \{chenxiqiu, yuetao, xuemeihu\}@nju.edu.cn}
}


\maketitle


\begin{abstract}
 High throughput video acquisition is a challenging problem and has been drawing increasing attention. Existing high throughput imaging systems splice hundreds of sub-images/videos into high throughput videos, suffering from extremely high system complexity. Alternatively, with pixel sizes reducing to sub-micrometer levels, integrating ultra-high throughput on a single chip is becoming feasible. Nevertheless, the readout and output transmission speed cannot keep pace with the increasing pixel numbers. To this end, this paper analyzes the strength of gradient cameras in fast readout and efficient representation, and proposes a low-bit gradient camera scheme based on existing technologies that can resolve the readout and transmission bottlenecks for high throughput video imaging. A multi-scale reconstruction CNN is proposed to reconstruct high-resolution images. Extensive experiments on both simulated and real data are conducted to demonstrate the promising quality and feasibility of the proposed method.
\end{abstract}


\input{introduction}
\input{related_work}

\input{gradient_based_acquisition}

\input{network}

\input{experiment}

\input{conclusion}


\bibliographystyle{plainnat}
\bibliographystyle{ieee_fullname}
\bibliography{gradcam}

\end{document}

%% file: introduction.tex
\section{Introduction}

High throughput imaging systems can provide detailed resolving capacity and a large field of view simultaneously, and are thus of great significance for many important applications, e.g., security monitoring, biological microscopy, and remote sensing. Several schemes for acquiring high throughput images/videos have been proposed in the past decade \cite{ben2010high, brady2012multiscale, fan2019video, cossairt2011gigapixel, ivezic2019lsst, kaiser2002pan}. However, these systems are typically bulky, complex, expensive, and power-consuming, hampering their practical applications in consumer-level areas.

The recent progress in developing sub-micron pixels \cite{samsung_small_pixel} makes the integration of ultra-high pixels on a single sensor possible. However, due to bottlenecks in readout speed and output transmission bandwidth, the scales of existing image sensors are limited to only hundreds of megapixels \cite{samsung, canon, sony104, sony150}.Regarding the bottleneck in readout speed, the single-slope analog-to-digital converter (SS-ADC) is most commonly adopted in commercial CMOS image sensors to read out signals \cite{totsuka20166}. Compared with other types of ADCs, it has significant advantages in terms of area and power consumption, which are crucial for consumer-level devices. However, the readout speed has become the bottleneck for high throughput image sensors; for instance, the frame rate of existing hundreds-of-megapixel sensors is limited to only several frames per second (fps) \cite{totsuka20166, samsung, canon, sony104, sony150}.As for the bottleneck in output transmission bandwidth, the most widely used output interface in commercial CMOS sensors has a bandwidth of only up to dozens of gigabits per second (Gbps). For example, the MIPI with a three-lane C-PHY v$2.0$ interface \cite{csi2} provides approximately 41.4 Gbps of bandwidth. However, to transmit an 8-bit high throughput video stream in real-time, a bandwidth of at least 240 Gbps is required.

To overcome these bottlenecks, we propose a low-bit gradient imaging scheme, enabling compact, low-cost, and low-power high throughput video acquisition. Based on the observation that low-bit image gradients can retain most high-frequency information, high-resolution (HR) images can be reconstructed with high fidelity using a low-bit high-resolution gradient (HRG) image and a low-resolution intensity (LRI) image. Since the low-bit HRG image can be read out through only a few comparison operations, the readout speed is significantly accelerated. Furthermore, because the low-bit HRG image has a minimal bit-depth and is highly sparse, it can be significantly compressed and transmitted within a limited bandwidth.
In this paper, we propose capturing a low-bit high throughput gradient image alongside a low-resolution (megapixel-level) image to achieve high throughput video acquisition. By combining low-bit gradient imaging with modified run-length and Huffman coding, we achieve readout speeds up to two orders of magnitude faster than conventional SS-ADCs and a data compression ratio below 10\%, enabling the real-time capture and transmission of low-bit high throughput gradient videos. To reconstruct the high throughput videos, we propose an end-to-end convolutional neural network (CNN) and demonstrate the effectiveness of our method through extensive experiments.
In particular, the contributions of this paper are as follows:
\begin{itemize}
\vspace{-2mm}
\item We \emph{propose} a gradient-based high throughput imaging scheme to realize compact, low-cost, and low-power high throughput video acquisition. Low-bit quantization of gradients is \emph{proposed} and \emph{explored} to achieve real-time high throughput readout.
\item We \emph{propose} a multi-scale fusion reconstruction CNN that effectively reconstructs the low-frequency information, structures, and details of scenes. Additionally, we \emph{develop} a modified run-length and Huffman coding module to ensure the data can be output using existing transmission interfaces.
\item We \emph{explore} the low bit-width settings for HRG and identify the gradient imaging scheme as the optimal choice for high throughput video imaging. Its effectiveness is further \emph{demonstrated} through both simulated and real-world experiments.
\vspace{-2mm}
\end{itemize}

%% file: related_work.tex
\vspace{-2mm}
\section{Related Work}
\vspace{-1mm}
\paragraph{Existing high throughput imaging systems}  The scanning based methods \cite{ben2010high} provide a simple scheme for high throughput imaging, at a cost of volume, weight, complexity and the video acquisition capability. To capture high throughput videos, Brady \etal \cite{brady2012multiscale} propose to use a multi-cameras system, with image stitching after capturing. Fan \etal \cite{fan2019video} extend this scheme to microscope for whole brain level real-time observation. Cossairt \etal \cite{cossairt2011gigapixel} present an architecture consisting of a ball lens shared by several small planar sensors. Besides, several survey telescopes \cite{ivezic2019lsst, kaiser2002pan} capture high throughput images/videos with the aid of sensor arrays. However, all these systems suffer from the large volume, heavy weight, and considerable complexity of the signal synchronization and capturing circuits.

\paragraph{Existing high throughput image sensors} Recently, several high-resolution CMOS image sensors have emerged. Totsuka \etal \cite{totsuka20166} present an APS-H-sized 0.25-billion-pixel high throughput CMOS image sensor. With the aid of improved column SS-ADC readout, the sensor can capture full-pixel images at 5 fps. Canon \cite{canon} uses 28 digital output channels to handle the enormous amount of data, which increases the complexity of the hardware design, achieving 9.4 fps for 0.122-billion-pixel high throughput images. Sony \cite{sony104,sony150} increases the readout speed through parallel signal processing, which arranges thousands of ADC circuits in a horizontal array to operate simultaneously, acquiring 0.15-billion-pixel high throughput images at 4 fps \cite{sony150} and 0.1-billion-pixel high throughput images at 6 fps \cite{sony104}. Samsung \cite{samsung} adopts a 3-stack fast readout sensor, enabling a readout speed of 10 fps for 0.108-billion-pixel high throughput images. In all, these CMOS image sensors with high integration density suffer from readout and transmission bottlenecks, lacking the capability to realize a video-rate high throughput sensor at present.

\paragraph{Gradient sensor/cameras based method} Recently, the gradient cameras \cite{tumblin2005want,Gottardi2009A,zhang2011gradient} are proposed for the advantage in fast frame rate, small data bandwidth and low power dissipation. Since the gradient images does not coincide with the human perceptions, Gallo \etal \cite{gallo2015retrieving} and Jayasuriya \etal \cite{2017CVPRW} propose to reconstruct the intensity images from the gradient measurements. However, for the absense of low frequency information, the recovered results suffer from severe reconstruction artifacts.

\paragraph{High-ratio image super-resolution (SR)} High-ratio image SR algorithms \cite{Shaham_2019_ICCV, lai2017deep, haris2018deep, guo2020closed} provide an alternative approach for reconstructing high throughput videos from megapixel-scale ones. However, due to the loss of high-frequency information, the ill-posedness of the SR problem becomes severe, especially under high-ratio super-resolution conditions (e.g., $> 16\times$ magnification). Consequently, the reconstructed images tend to be overly smooth and fine details cannot be recovered, which prevents the reconstruction of high-resolution images with high fidelity and reliability.

\paragraph{Reference-based image super-resolution (RefSR)} In the multi-camera setting, HR image can be reconstructed by a LR image and a reference HR image from other viewpoints or frames \cite{zheng2018crossnet, cheng2020dual}. Cheng \etal \cite{cheng2020dual} propose to reconstruct HR videos with LR videos and HR video of lower frames, in which way the transmission bandwidth could be reduced a lot through only transferring HR low frame rate videos and LR high frame rate videos. Zheng \etal \cite{zheng2018crossnet} propose to reconstruct the HR image of a certain viewpoint of camera with reference from the HR image of another viewpoint of camera, in which way a high resolution image that contains a large  range of viewpoints could be reconstructed with only capturing the low resolution multi-view LR image and a reference HR image of a certain view point and the data transmission bandwidth could also be reduced a lot. 
However, to utilize these ideas in reducing the data transmission bandwidth for high throughput videos, the interpolation ratio of framerate (i.e. the ratio between the frame rate of the LR video and that of HR video) or viewpoints become too high to guarantee reconstruction with high fidelity. 

In this paper, we propose a low-bit gradient imaging scheme, which greatly speed up the readout and reduce the transmission bandwidth, promising for realizing a compact, inexpensive and low-power high throughput video camera.

%% file: gradient_based_acquisition.tex

\section{Low Bit Gradient Based Imaging Strategy}
As mentioned above, the major issue in realizing a single-chip high throughput video sensor is the bottleneck in readout and transmission speed. Inspired by the fact that a low-bit gradient map can still preserve high-frequency information effectively, and that low-frequency information can be captured with a low-resolution image, we propose capturing a low-bit HRG map alongside an LRI image to realize compact, low-cost, and low-power high throughput video cameras. Instead of capturing high-resolution images directly, the low-bit HRG-based imaging scheme significantly reduces the requirements for readout speed and transmission bandwidth, enabling the acquisition of real-time high throughput videos using existing image sensor technologies.

\subsection{Intuitive analysis}

\begin{figure}[htb]
\centering
\includegraphics[width=\linewidth]{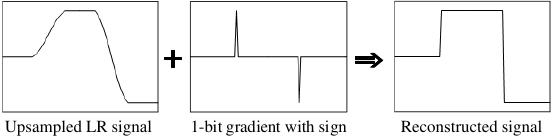}
\caption{The 1D diagram of the principle of proposed low bit gradient based scheme. Note that the LR signal is up-sampled, so that the signal has obviously blurry effect.}
\label{fig: one_dim}
\end{figure}

The proposed scheme is composed of two input, i.e., the low bit HRG and the LRI. In Fig. \ref{fig: one_dim}, a one-dimensional (1D) signal diagram of the proposed method is illustrated to give an intuitive understanding. The low bit gradient signal indicates the exact location of the steps, and combined with the low frequency, the sharp HR signal can be exactly recovered.

\begin{figure}[htb]
\centering
\includegraphics[width=\linewidth]{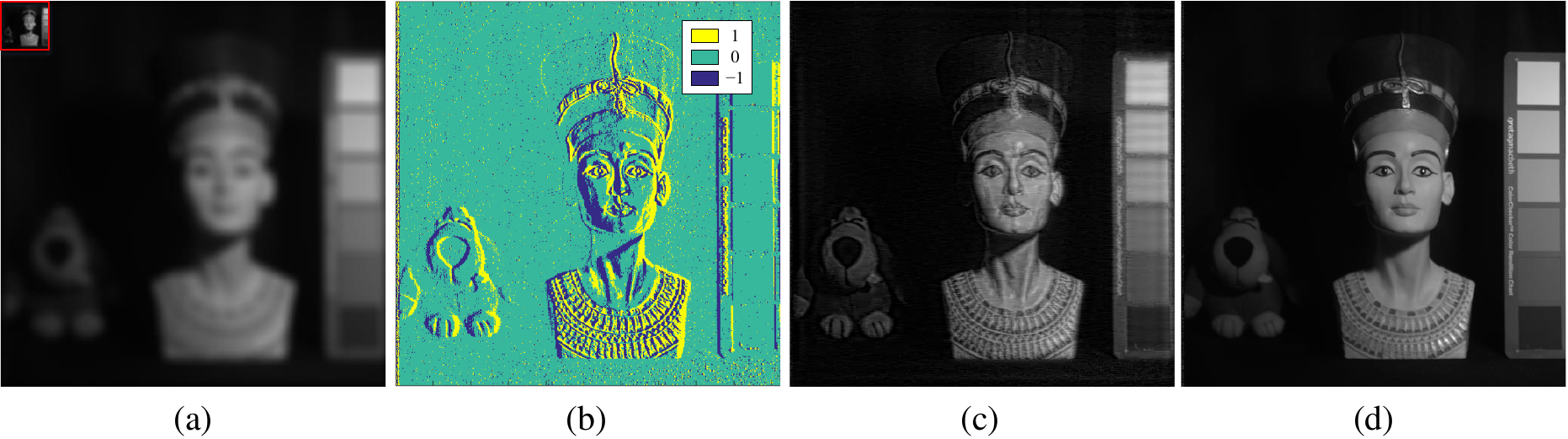}
\caption{The principle verification of the proposed method with a simple reconstruction method, (a) the LRI image (up-left in the red box) and its 8$\times$ up-sampled image, (b) the HRG map, (c) the reconstructed HR image by Eq.~\eqref{eq: simple_solve}, and (d) the ground truth.}
\label{fig: simple}
\end{figure}

To demonstrate the effectiveness of proposed scheme for 2D image cases, we introduce a simple reconstruction method by solving the following problem,
\begin{equation}
  \min_{\bf I} ||{\bf L}\uparrow - {\bf I}||^2_2 + \lambda ||\nabla_x{\bf I} - {\bf G}||^2_2 + \beta ||{\bf I}||^2_2,
  \label{eq: simple_obj}
\end{equation}
where ${\bf I}$ is the latent HR image, ${\bf L}$ is the LRI image, $\uparrow$ denotes the $8\times$ up-sampling operator, ${\bf G}$ is the low bit HRG map, $\nabla_x$ is the differential operation in the row direction, $\lambda$ and $\beta$ are the weights of the gradient constraint and the L-2 norm regularization terms respectively. Eq.~\eqref{eq: simple_obj} can be solved in closed form 
\begin{equation}
  {\bf I}=\mathcal{F}^{-1}\{\frac{\mathcal{F}(\uparrow)^*\odot\mathcal{F}(\bf L) + \lambda \mathcal{F}(\nabla_x)^*\odot\mathcal{F}(\bf G)}{\mathcal{F}(\uparrow)^*\odot\mathcal{F}(\uparrow) + \lambda\mathcal{F}(\nabla_x)^*\odot\mathcal{F}(\nabla_x) + \beta} \},
  \label{eq: simple_solve}
\end{equation}
where $\mathcal{F}$ and $\mathcal{F}^{-1}$ are the Fourier and inverse Fourier transforms respectively, $(\cdot)^*$ is the conjugate operation, $\odot$ denotes point-wise multiplication and note that the division also denotes the point-wise division. As shown in Fig.~\ref{fig: simple}, the result of the closed form reconstruction method contains almost all the high frequency details, and thus showing the potential of proposed low bit gradient based imaging schemes for high quality reconstruction. In the following, we will introduce our neural network based reconstruction module for the HR reconstruction with the proposed acquisition scheme, and demonstrate the superiorities of proposed schemes.

\subsection{Hardware feasibility}
We present a feasible system blueprint for the propose imaging strategy. The system cores are two image sensors, i.e., a common LRI image sensor and an HRG sensor. A beam splitter is introduced to sense the same image formed by the objective lens. The LRI image sensor can be dozens of times (8$\times$8 times in our paper) smaller than the reconstructed images, so that a consumer-level 6K$\times$4K sensor, e.g., \cite{sony}, can be used. The HRG sensors has exactly the same pixel array architecture of the common CMOS image sensor technologies, except the readout and transmission part. The readout process can be achieved by a simple comparator with only a few compare operations, which therefore can be readout much faster and conserve more energy. Compared with the common SS-ADCs used in CIS sensors, the proposed method can be orders of magnitude faster. After a simple run-length and entropy coding, the data can be easily compressed, and the transmission bandwidth can be easily meet by the widely used camera interface, i.e., MIPI CSI-2 \cite{csi2} with maximum transmission bandwidth of up to 41.4 Gbps.

In fact, a simple prototype gradient camera~\cite{Gottardi2009A} is implemented, verifing its superiorities on high speed readout, low transmission bandwidth and low energy consumption. Except the gradient sensor, all the techniques required in the proposed scheme are off-the-shelf, greatly reducing the difficulties and risks for implementing the proposed scheme. 


%% file: network.tex
\section{Approach}

\begin{figure*}[htb]
\centering
\includegraphics[width=\linewidth]{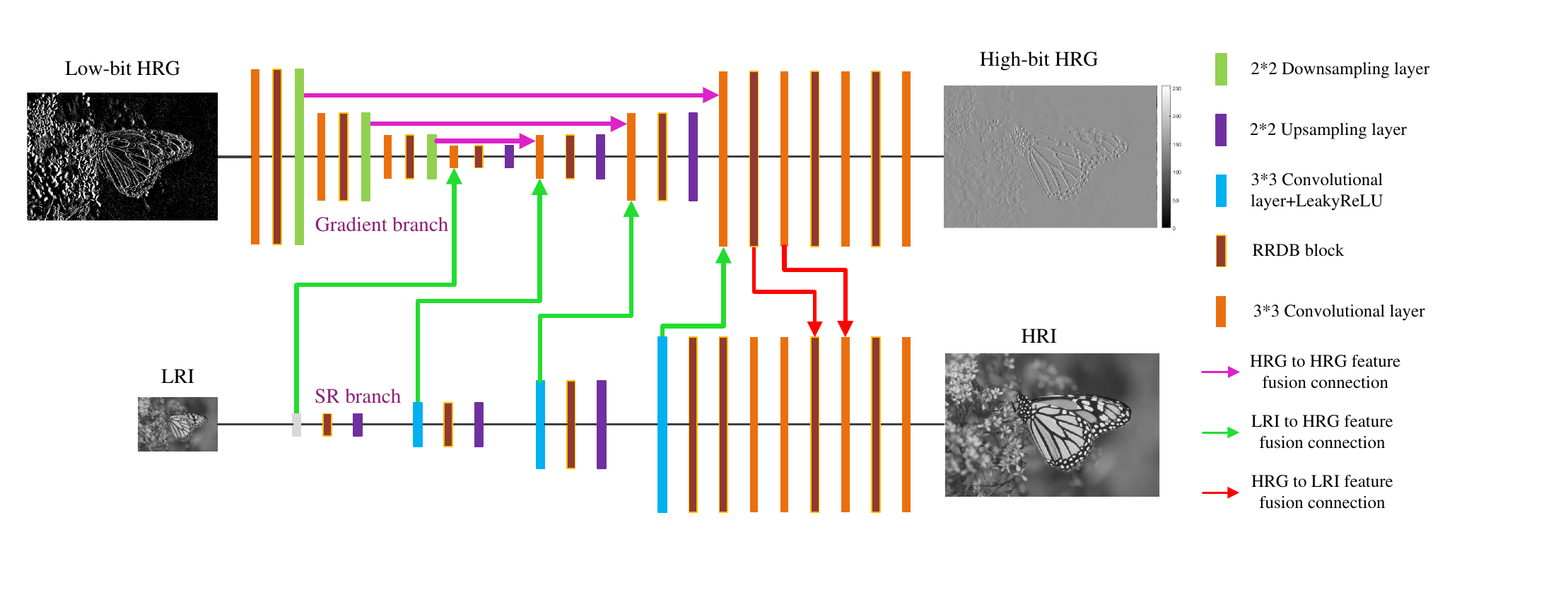}
\caption{The diagram of our multi-scale fusion reconstruction network. The network takes low-bit HRG map and LRI image as input, and outputs the reconstructed HR image and high-precision HRG map in two directions.}
	\label{fig: network}
\end{figure*}

In this section, we proposed the method to generate the high quantliy high throughput images using the multi-scale fusion reconstruction network(MSFRnet) from low bit high resolution gradient map and low resolution intensity images. The structure details of three branches, i.e., the SR branch and two Gradient branches, and the feature fusion connection between the branches will be introduced in the following.

\subsection{Overview}

Because of the scale gap between HRG map and LRI image, we design a multi-scale feature fusion CNN-based reconstruction network, which can fuse the high frequency information of HRG map and the low frequency information of LRI image at different scales.
The network can be divided into three branches according to the function of different branches, SR branch and two gradient branches.
Fig.~\ref{fig: network} shows the diagram of MSFRnet.
The SR branch takes LRI image as input and generates the correspponding 8$\times$8 super resolved image. In the second half of SR branch, it fuses the features of two gradient branches to help the reconstruction of HR image.
Specifically, the HRG map contains high frequency information and the LRI image contains low frequency information, which are complementary for the HR image reconstruction.
The gradient branches takes low-bit HRG map as input and outputs high-bit HRG map in the corresponding direction. At the same time, it fuses the features of SR branch with multi-scales to guide the reconstruction of high-bit HRG map. 
Details of the network are introduced below.

\subsection{Details in Architecture}

\paragraph{SR branch}
SR branch can be divided into two parts. The first part completes the 8$\times$8 super resolution of LRI image and obtains the features at various scales, which can be used to guide the reconstruction of gradient branches. The second part fuses the gradient features from gradient branches to complete the final HR image reconstruction. As shown in Fig.~\ref{fig: network}, the SR branch adopts a progressive super resolution method, that is to say, we use three times of 2$\times$2 upsampling layers to achieve 8$\times$8 super resolution. The advantage of this method is that it can obtain features in different scales and can be used to guide the reconstruction of gradient branches. In the second half of the network, SR branch combines the features of two gradients, that is, it contains the gradient information of two directions, and finally outputs clear images.

\paragraph{Gradient branches}
Due to the loss of accuracy caused by low-bit HRG map, we design two gradient branches to restore the high-bit HRG map in two directions. The two gradient branches have the same structure, but the supervision map is different, which are high-bit HRG map in $x$ and $y$ directions. The HRG map for an image $I$ can be obtained by computing the difference between adjacent pixels:
\begin{eqnarray}
G_x(x) &=& I(x+1, y)-I(x, y), \\
G_y(x) &=& I(x, y+1)-I(x, y),
\end{eqnarray}
The first half of gradient branch adopts the structure similar to U-net~\cite{ronneberger2015u}, but different from U-net, we use Residual-in-Residual Dense Block(RRDB) proposed in \cite{wang2018esrgan} as the basic feature extraction block instead of simple convolutional block. We alse use max pooling layer as down-sampling layer and deconvolutional layer as up-sampling layer. We combine RRDB and max pooling layer as a basic 2$\times$2 down-sampling module. In the downsampling process, we use three times of 2$\times$2 down-sampling module to obtain four different scale features (including the original scale), which will be fused with the features of up-sampling module through concat operation. We combine RRDB and deconvolutional layer as a basic 2$\times$2 up-sampling module, symmetrically, we use three times of 2$\times$2 up-sampling layers to restore the features to the original scale. In the process of up-sampling, the features of the SR branch are fused to the gradient branch, this is because the former contains low-frequency imformation of the image, which can help the reconstruction of high-bit HRG map.

\paragraph{Feature fusion connection}
Intuitively, the low frequency features of SR branch can guide the reconstruction of high-bit HRG map, and the high frequency features of gradient branches can also guide the reconstruction of HR image. Therefore, we use multiple feature concat operations to fuse low frequency features and high frequency features in MSFRnet. As shown in Fig.~\ref{fig: network}, MSFRnet has three different types of fuse connections: HRG to HRG, LRI to HRG and HRG to LRI feature fuse connections, which are located in different parts of the network and play different roles.
Specifically, in the first half of the whole network, all three branches are a kind of multi-scale structure. In this part, we fuse the low frequency features of SR branch into the second half of U-shape network in gradient branches to guide the reconstruction of high precision HRG map, and we also fuse the high frequency features of gradient branch in different scales, which is same as U-net. In the second half of the whole network, we fuse the high frequency features of radient branches into SR branch to complete the final HR reconstruction.
Except for the layers with feature fusion connection, the number of input and output feature maps of each layer is $16$. If the input contains feature fusion connection, the number of input feature maps is a multiple of $16$, which is determined by the number of feature fusion connections, e.g. $32$, $48$.

\paragraph{Loss function}
For network training, we adopt the mean squared error between the ground truth HR image and the reconstructed result as the loss function of SR branch. We also adopt the mean squared error between the low-bit(LB) HRG map and high-bit(HB) HRG map as the loss function of two gradient branches, which correspond to $x$ and $y$ directions respectively.
\begin{eqnarray}
\mathcal{L} &=& \mathcal{L}_{SR}+\mathcal{L}_{G_x}+\mathcal{L}_{G_y} \notag \\ 
&=& \|\mathcal{F}(I^{LR})-I^{HR}\|_2+\beta\|\mathcal{F}(G^{LB}_{x})-G^{HB}_x\|_2 \notag \\ 
&&+\gamma\|\mathcal{F}(G^{LB}_{y})-G^{HB}_y\|_2
\label{eq: loss_func}
\end{eqnarray}

\subsection{Quantization and Coding Strategy}
\label{sec:quantization}

In this section, we discuss the quantization and coding strategy in detail. To achieve a comprehensive balance among reconstruction quality, readout speed, and transmission bandwidth, we introduce a generalized nonlinear quantization scheme with $L = 2^n + 1$ discrete levels. This scheme is efficiently implemented by using comparators to perform multi-level shifted comparisons on intensity measurements $I_a$ and $I_b$ from adjacent photodiode columns. Mathematically, the quantization procedure can be represented by a multi-threshold function:
\begin{equation}
  Q(g_{ab}) = \text{quantize}(I_a - I_b, \{\tau_i\}),
  \label{eq:gen_quant}
\end{equation}
where $\{\tau_i\}$ denotes a set of thresholds determining the quantization intervals. The selection of these thresholds directly influences the trade-off between the required transmission bandwidth and the reconstruction fidelity.

To utilize the sparsity of the quantized gradient map and further compress the data, a modified run-length coding (RLC) and entropy coding module is introduced following the quantization. Instead of directly outputting each value, the number of repetitions and the corresponding levels are recorded using a counter sequence and a value sequence, respectively. For the value sequence, we propose a state-reduction strategy to minimize the bit depth. In a general $2^n+1$ quantization scheme, representing each level would typically require $n+1$ bits. However, since RLC only records values when a transition occurs, the next value is restricted to the remaining $2^n$ possible states. This allows the transition to be uniquely identified using only $n$ bits, effectively saving one bit per transition compared to naive encoding. Furthermore, the counter sequence is optimized using 8-bit segments. When the number of repetitions is greater than or equal to 255, a value of 255 is recorded and the remainder is stored in the subsequent segment, allowing the decoder to reconstruct the total length by summation. Finally, as the distribution of the repetition numbers is highly non-uniform, Huffman coding is applied to the counter sequence for further compression.

%% file: experiment.tex
\section{Experiments}
\label{sec:exp}
In this section, we first explore the gradient imaging scheme with different low bit-width setting.  
To demonstrate the effectiveness of the proposed method, we compare with the alternative imaging scheme that could be adopt for realizing high throughput video imaging, containing high ratio superresolution imaging method~\cite{shi2016real, ledig2017photo, lai2017deep, tong2017image, lim2017enhanced, haris2018deep, zhang2018image, dai2019second, guo2020closed} and reference based superresolution methods~\cite{zheng2018crossnet, cheng2020dual}. Noise robustness and ablation study of the neural network is implemented to further validate the proposed neural network. 

With the superiority demonstrated compared with the other methods, we further demonstrate the effectiveness of the proposed method on high throughput image containing both fidelity and compression validation on the high throughput dataset~\cite{Wang_2020_CVPR} and qualitative results on real captured data.

\subsection{Exploration of Gradient Imaging Scheme}

Tumblin \etal \cite{tumblin2005want} propose to simultaneously collect gradient values in the $x$ and $y$ direction, i.e. two gradient values are captured for each pixel. However, the amount of data in the gradient domain becomes twice as much as that in the spatial domain. Since the transmission bandwidth imposes strict limit on the amount of data, how to choose a more effective gradient acquisition scheme is very important in the task of real-time high throughput imaging. In this section, we explore the low bit setting for the gradient imaging scheme, 5 different types ranging from 1 to 2 bit with with different bit width allocation and gradient directions, and discuss the feasibility for realizing realtime high throughput imaging in terms of readout speed, transmission bandwidth, and image reconstruction fidelity. 

\begin{figure*}[htb]
\centering
\includegraphics[width=\linewidth]{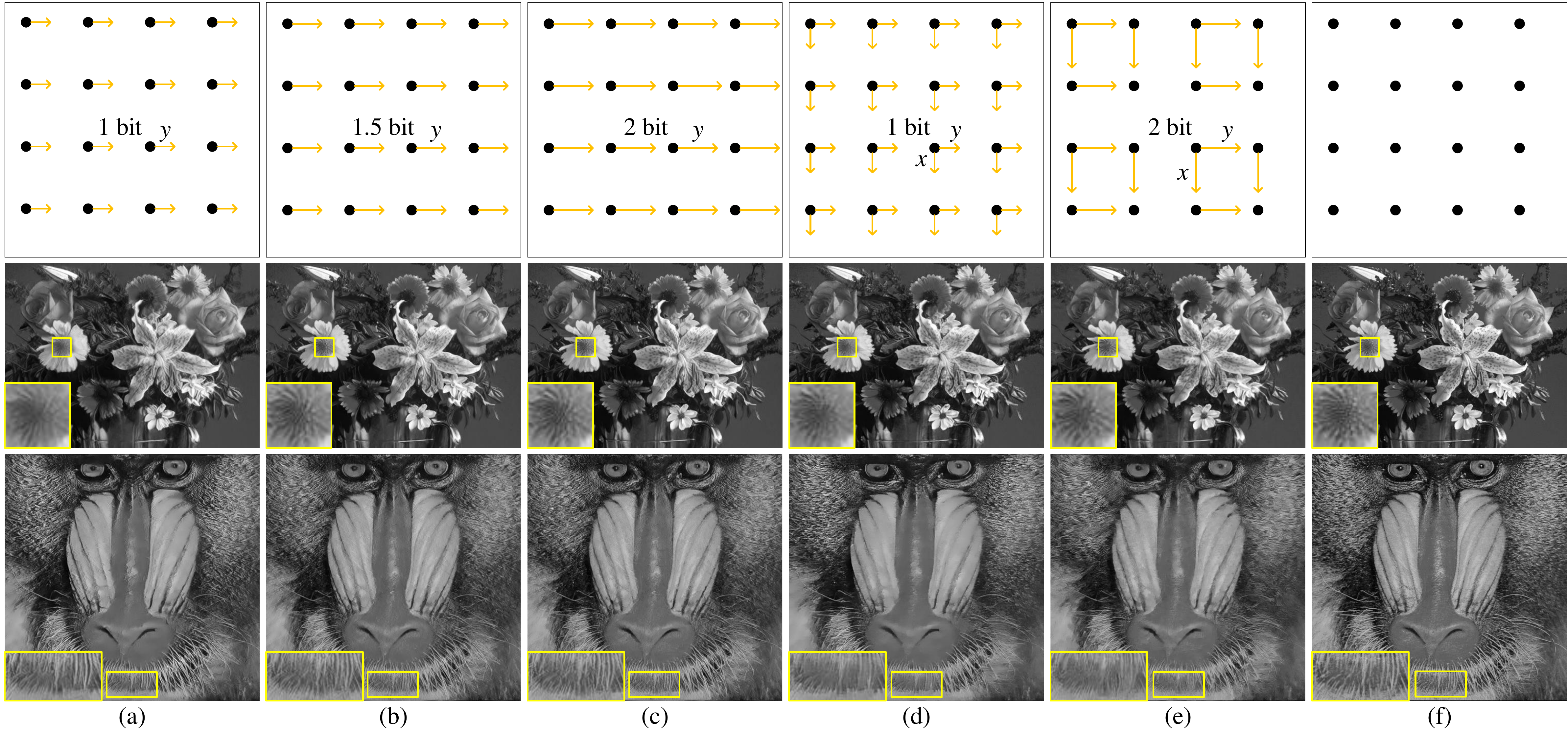}
\caption{The reconstruction results of different HRG acquisition schemes, the first row is the schematic diagram of HRG gradient acquisition scheme. The direction of the arrow represents the direction of the gradient values. There are two directions: $x$ and $y$, the length of the arrow represents the bit number of the HRG gradient map. There are three kinds of bit numbers, from short to long, representing $1$ bit, $1.5$ bits and $2$ bits respectively. (a). $1$ bit, $1$ direction, (b). $1.5$ bits, $1$ direction, (c). $2$ bits, $1$ direction, (d). $1$ bit, $2$ directions, (e). $2$ bits, $2$ directions, but each pixel value in each direction will only be calculated once gradient, (f). groundtruth.}
\label{fig: gradient_structure}
\end{figure*}

\begin{table}[htbp]
\centering
\caption{Experimental results with different gradient acquisition scheme.}
	\scalebox{0.85}{
	\begin{tabular}{c|cccccc}
	\hline
{\multirow{2}[0]{*}{Scheme}} &
				\multicolumn{1}{c}{bits} & \multicolumn{1}{c}{1}& \multicolumn{1}{c}{1.5} & \multicolumn{1}{c}{2} & \multicolumn{1}{c}{1} & \multicolumn{1}{c}{2}\\

				& \multicolumn{1}{c}{directions} & \multicolumn{1}{c}{1} & \multicolumn{1}{c}{1} & \multicolumn{1}{c}{1} & \multicolumn{1}{c}{2} & \multicolumn{1}{c}{2}\\
	\hline
{\multirow{2}[0]{*}{PSNR}} 
				& \multicolumn{1}{c}{Set5 \cite{set5}} & \multicolumn{1}{c}{28.97} & \multicolumn{1}{c}{31.78} & \multicolumn{1}{c}{32.31} & \multicolumn{1}{c}{31.57} & \multicolumn{1}{c}{31.39}\\
				& \multicolumn{1}{c}{Set14 \cite{set14}} & \multicolumn{1}{c}{26.62} & \multicolumn{1}{c}{29.05} & \multicolumn{1}{c}{29.21} & \multicolumn{1}{c}{29.01} & \multicolumn{1}{c}{28.21}\\
	\hline
{\multirow{2}[0]{*}{SSIM}} 

				& \multicolumn{1}{c}{Set5 \cite{set5}} & \multicolumn{1}{c}{0.875} & \multicolumn{1}{c}{0.929} & \multicolumn{1}{c}{0.945} & \multicolumn{1}{c}{0.922} & \multicolumn{1}{c}{0.924}\\
				& \multicolumn{1}{c}{Set14 \cite{set14}} & \multicolumn{1}{c}{0.832} & \multicolumn{1}{c}{0.896} & \multicolumn{1}{c}{0.904} & \multicolumn{1}{c}{0.900} & \multicolumn{1}{c}{0.871}\\
	\hline
{\multirow{1}[0]{*}{TB}} &
				\multicolumn{1}{c}{-} & \multicolumn{1}{c}{0.125}& \multicolumn{1}{c}{0.1875} & \multicolumn{1}{c}{0.250} & \multicolumn{1}{c}{0.250} & \multicolumn{1}{c}{0.250}\\
	\hline
	{\multirow{1}[0]{*}{RS}} &
				\multicolumn{1}{c}{-} & \multicolumn{1}{c}{256}& \multicolumn{1}{c}{128} & \multicolumn{1}{c}{85} & \multicolumn{1}{c}{128} & \multicolumn{1}{c}{85}\\
	\hline
	\end{tabular}
			}
	\label{tab:dif_gradient_scheme}
\end{table}

To train the proposed network with different gradient imaging scheme and compare the reconstruction fidelity, we adopt the training dataset DIV2K~\cite{timofte2017ntire}.
We adopt MSE loss and the Adam optimizer \cite{kingma2014adam} with a  initial learning rate of $1\times 10^{-4}$, and we set $\beta$ and $\gamma$ to $0.1$ in Eq.~\eqref{eq: loss_func}. The learning rate is multiplied by $0.5$ for every $200$k iterations during the training. Kernel size of the convolutional layers are set to $3$ and the number of feature maps is set to $16$. All the experiments are implemented by PyTorch on NVIDIA RTX 2080Ti GPUs.

As shown in the first row of Fig.~\ref{fig: gradient_structure}, we compare 5 possible gradient acquisition schemes with average bit width not larger than 2. As shown, taking a $4\times4$ pixel array as example, 
for gradient imaging with only one direction, we design three kind of possible bit width setting, i.e. 1, 1.5 and 2. The gradient is quantized into \{0, 1\}, \{-1, 0, 1\}, \{-2, -1, 0, 1\}, respectively. 
For gradient imaging with two directions (i.e. both x and y direction), to realize the low bit average bit-with, we design two types of gradient imaging scheme, 1 bit for two directions and 2 bit for two directions while with half spatial resolution. 

We trained the proposed reconstruction network with different gradient scheme with the same 600K iters and the reconstruction fidelity are shown in Table \ref{tab:dif_gradient_scheme}. As shown, for gradient imaging with only one direction, the larger the bit width, the higher the reconstruction quality. When the bit width is larger than 1 bit (i.e. 1.5 and 2 bit), the reconstruction quality is high with PSNR larger than 30 dB and SSIM larger than 0.9. As for the gradient imaging scheme with two directions, the reconstruction are also of high quality, however, the reconstruction quality is not better than the second gradient scheme with 1 direction 1.5 bit gradient. The comparison between the second gradient scheme and the fourth gradient scheme shows that the sign of gradient plays an essential role in image retrieval, which is more important than the average bit width. The comparison between the third and fifth gradient imaging scheme shows that it is important to keep the gradient with full resolution. To visualize the difference of different gradient imaging scheme, we further show the reconstructed result of different gradient setting. As shown in Fig.~\ref{fig: gradient_structure}, with different setting of gradient scheme, the detail preserving ability is different, which is consistent with the quantitative comparisons. Through quantitative and qualitative comparisons, we could find out that, in terms of reconstruction quality with low bit gradient scheme, the ternary and quaternary gradient imaging scheme with only 1 direction is better in preserving the details such as the center of the stamen, the fringe of the tiara, and the moustache of lion.

As shown in Table \ref{tab:dif_gradient_scheme}, the ratio of transmission bandwidth (TB) for different gradient schemes compared with the 8 bit original image is 0.125, 0.1875, 0.25 respectively. For transmission of 8 bit high throughput video images at bandwidth 41.4 Gbit/s (MIPI CSI-2 \cite{csi2}), the corresponding frame rate is 41, 27, 20, 20 and 20 fps for different scheme.

As for readout speedup (RS), for 8 bit binary quantization, 256 times of comparison is required. However, current technologies has been limited by the read out speed of around 10 Gbit/s~\cite{totsuka20166,canon,sony150,samsung}.
Instead of reading out the pixel value directly, readout the quantized gradient could greatly improve the readout speed. For quantization with 1, 1.5 and 2 bit width, only 1, 2, 3 comparisons are required respectively, which could theoretically improve the readout speed by around 256, 128, 85 times. With the readout speed increased by more than 85 times, the corresponding read out speed is no longer the limit for real time high throughput video imaging.

In terms of hardware implementation feasibility, since with the mature framework of CMOS sensor technology, the readout is realized through putting the ADCs at the end of each column and a set of addressing circuit. To realize the gradient quantization in only one direction, we only need to replace the ADC with comparators. However, for implementing gradient acquisition of two directions, we need to redesign the address circuit for readout, which introduce more burden and power consumption to the sensor.

In all, through thoroughly consideration of both reconstruction quality, required transmission bandwidth, readout speedup and hardware implementation feasibility, we choose the second scheme, i.e. ternary gradient imaging scheme, for real time high throughput video imaging. 

\begin{figure}[htbp]
\centering
\includegraphics[width=0.5\textwidth]{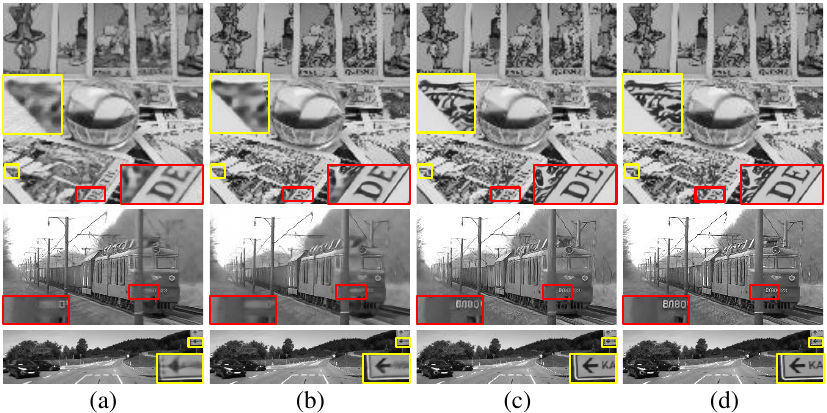}
\caption{Experiental results on different datasets. top: Stanford Light field, middle: YUP++, bottom: KITTI. (a). CrossNet, (b). AWnet, (c). Ours, (d). groundtruth.}
\label{fig: compare_RefSR}
\end{figure}

\subsection{Effectiveness demonstration on datasets}

\begin{figure}[htbp]
\centering
\includegraphics[width=0.5\textwidth]{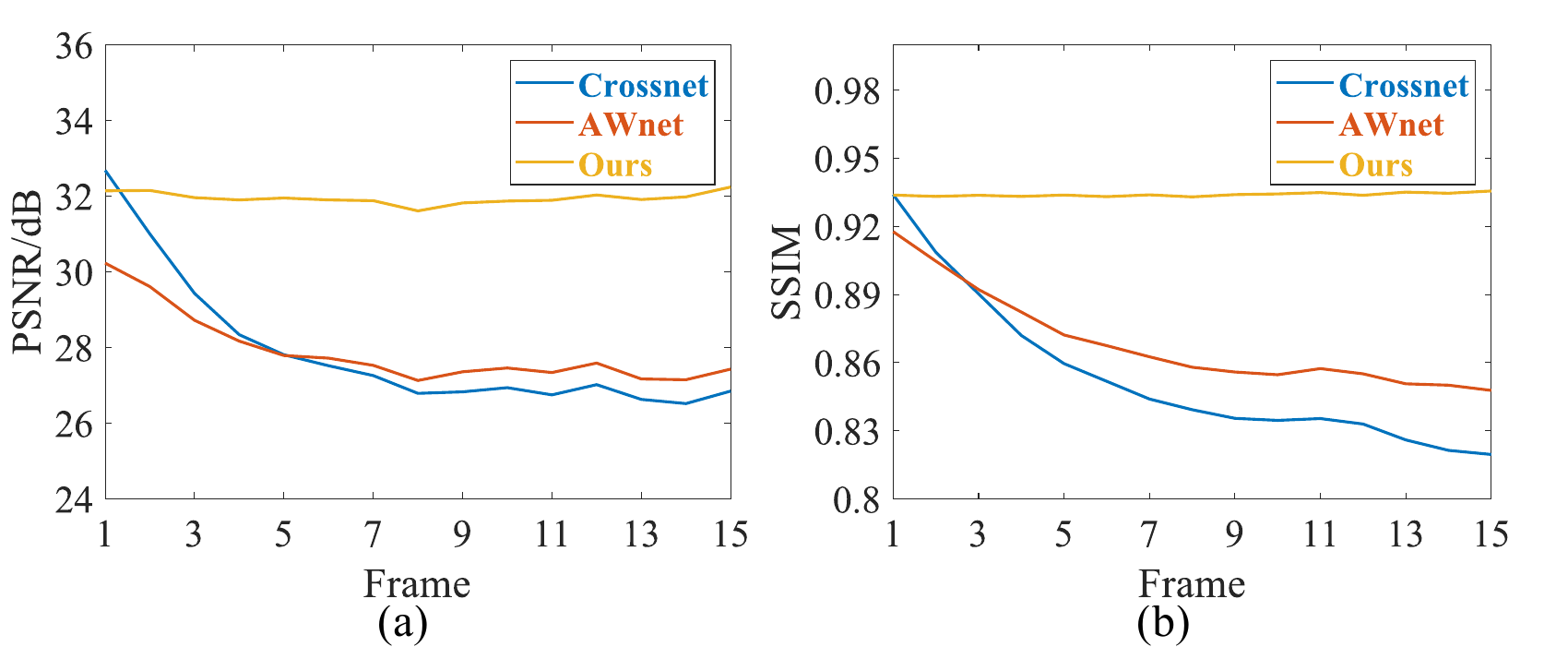}
\caption{Influence of reference HR image on YUP++ dataset. (a). PSNR w.r.t frame, (b). PSNR  w.r.t ratio of reference HR image and LR image.}
\label{fig: yup_psnr}
\end{figure}

\paragraph{Compare with RefSR methods}
Recently, several RefSR algorithms have been proposed ~\cite{zheng2018crossnet, cheng2020dual}, which could also be utilized for reducing the data transmission bandwidth and realize real time high throughput imaging. In order to compare with these methods, we tested these methods ~\cite{zheng2018crossnet, cheng2020dual} in public accessible videos and light field datasets (e.g. KITTI ~\cite{menze2015object}, YUP++ ~\cite{feichtenhofer2017temporal} and Stanford Light field~\cite{stanfordLF} datasets). Since the authors do not provide the pretrained models, we retrained CrossNet and AWnet following the training stategy suggested in ~\cite{zheng2018crossnet, cheng2020dual} and use the same trainning dataset as ~\cite{cheng2020dual}. The quantitative results comparison are shown in Table \ref{tab:RefSR} and the visual result from these datasets are shown in Fig.~\ref{fig: compare_RefSR}. KITTI is a two-view video dataset, which has $54$cm baseline distance for two cameras. In the KITTI dataset, for RefSR, we use the HR right view image as the input reference HR image and LR left view image as the input LR image to reconstruct super-resolved left view image.  At the same point, in our method, we use HRG map and LRI image from left view as input. Because LR image and reference HR image have different views, some regions in LR image cannot find corresponding HR regions in reference HR image, which leads to some artifacts in the reconstructed image. YUP++ ~\cite{feichtenhofer2017temporal} is a video dataset of dynamic scenes, the frame rate can only reach $1$ fps for high throughput video, so for RefSR methods, we need to reconstruct SR images with only $1$ reference HR frame. We use $1$ HR frame as the reference HR image and next $15$ LR frames to reconstruct the SR video, the average results are shown in Table ~\ref{tab:RefSR}.
In order to study the influence of the difference between reference HR image and LR image on the reconstruction results, we also tested the results at different frame distance of reference HR frames and LR frames. As shown in Fig.~\ref{fig: yup_psnr}, when the reference HR frame is farther and farther away from the LR frame, the reconstruction result of RefSR methods will drop rapidly, same results are reported in \cite{cheng2020dual}. For a high throughput video, the existing technology can only capture a very low frame rate reference HR video sequences, e.g., $1$ fps. In this case, reference-based SR methods cannot restore high quality video. The Stanford light field dataset contains $17\times17$ angular samples (grid), from which we set HR image at $(0, 0)$ as the reference image and reconstruct the LR images at $(0, 1)$ to $(0, 16)$, the average results are shown in Table~\ref{tab:RefSR}. Fig.~\ref{fig: compare_RefSR} shows visual result in three different datasets, we can see there are many artifacts in the reconstruction results of RefSR methods. To sum up, our method outperforms the exist RefSR methods, and is not limited by frame rate or view angle.

\begin{table}[htbp]
\centering
\caption{Performance comparison with state-of-the-art algorithms for 8$\times$ RefSR.}
	\scalebox{0.9}{
	\begin{tabular}{c|ccc}
	\hline
	\multicolumn{1}{c|}{\multirow{2}[0]{*}{Algorithms}} &
				\multicolumn{1}{c}{CrossNet} & \multicolumn{1}{c}{AWnet} & \multicolumn{1}{c}{Ours} \\
				& \multicolumn{1}{c}{PSNR/SSIM} & \multicolumn{1}{c}{PSNR/SSIM} & \multicolumn{1}{c}{PSNR/SSIM}\\
    \hline
	KITTI  & 24.92/0.798    &   26.01/0.836    &   29.71/0.917 \\
	\hline
	StanforfLF  & 32.68/0.933    &   34.92/0.954    &   35.64/0.959 \\
	\hline
	YUP++ & 27.91/0.854    &   27.90/0.869    &   31.95/0.934 \\
	\hline
	\end{tabular}
				}
	\label{tab:RefSR}
\end{table}

\paragraph{Influence of noise}

\begin{figure}[htbp]
\centering
\includegraphics[width=0.5\textwidth]{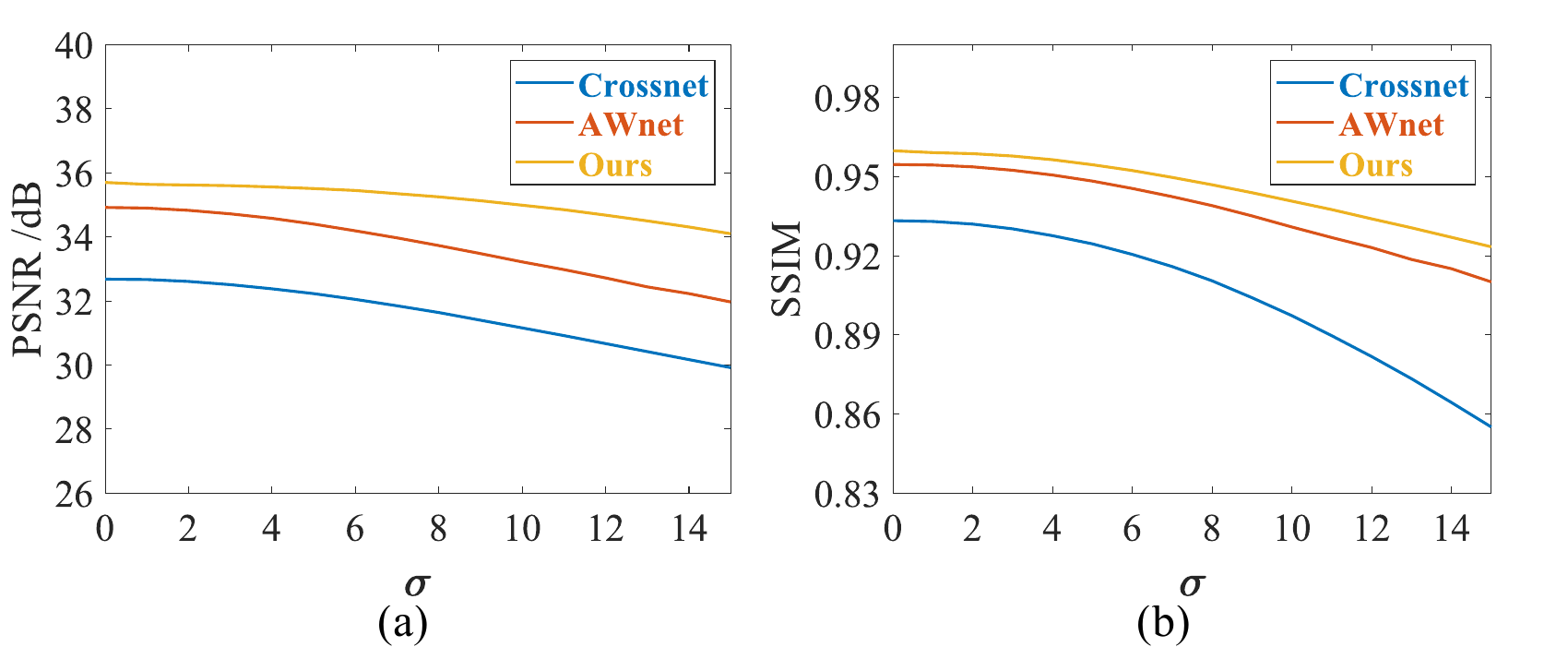}
\caption{Experimental results of synthesizing noise images on Stanford Light field dataset.}
\label{fig: noise_psnr_ssim}
\end{figure}

Fig.~\ref{fig: noise_psnr_ssim} shows the experimental results at different synthetic noise levels on the Stanford Light Field dataset. Compared with HR images, the noise of LRI images could be ignored. In our experiment, the input is a clear LRI image and a HRG map with different levels of noise, meanwhile, for RefSR methods, the input is a clear LRI image from $(0,1)$ to $(0,16)$ and a HR reference image at $($0$,$0$)$ with different levels of noise. Specifically, the bit width of the groundtruth image is 8 bit and $\sigma$ denotes the standard deviation of the added Gaussian white noise. Fig.~\ref{fig: noise_psnr_ssim} shows the average results from $(0,1)$ to $(0,16)$ with different methods. The models we use for testing are all trained on non-noise datasets, as shown, our method is more robust to noise compared to the other two method.

\paragraph{High resolution image reconstruction}

\begin{figure}[htbp]
\centering
\includegraphics[width=0.5\textwidth]{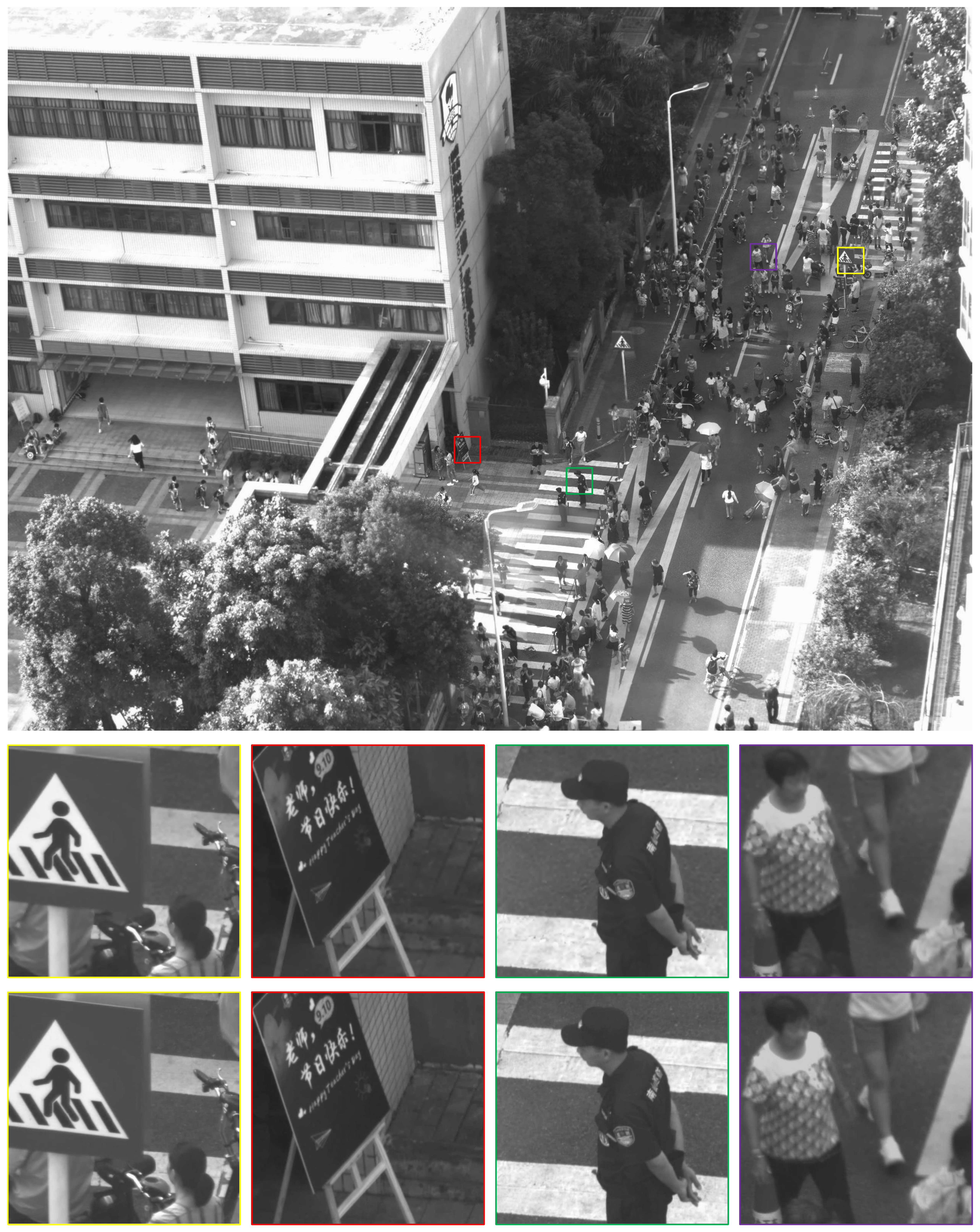}
\caption{Test results on High resolution images. Groundtruth: top, reconstruction result: bottom.}
\label{fig: gigapixels}
\end{figure}

\begin{figure*}[htbp]
\centering
\includegraphics[width=\textwidth]{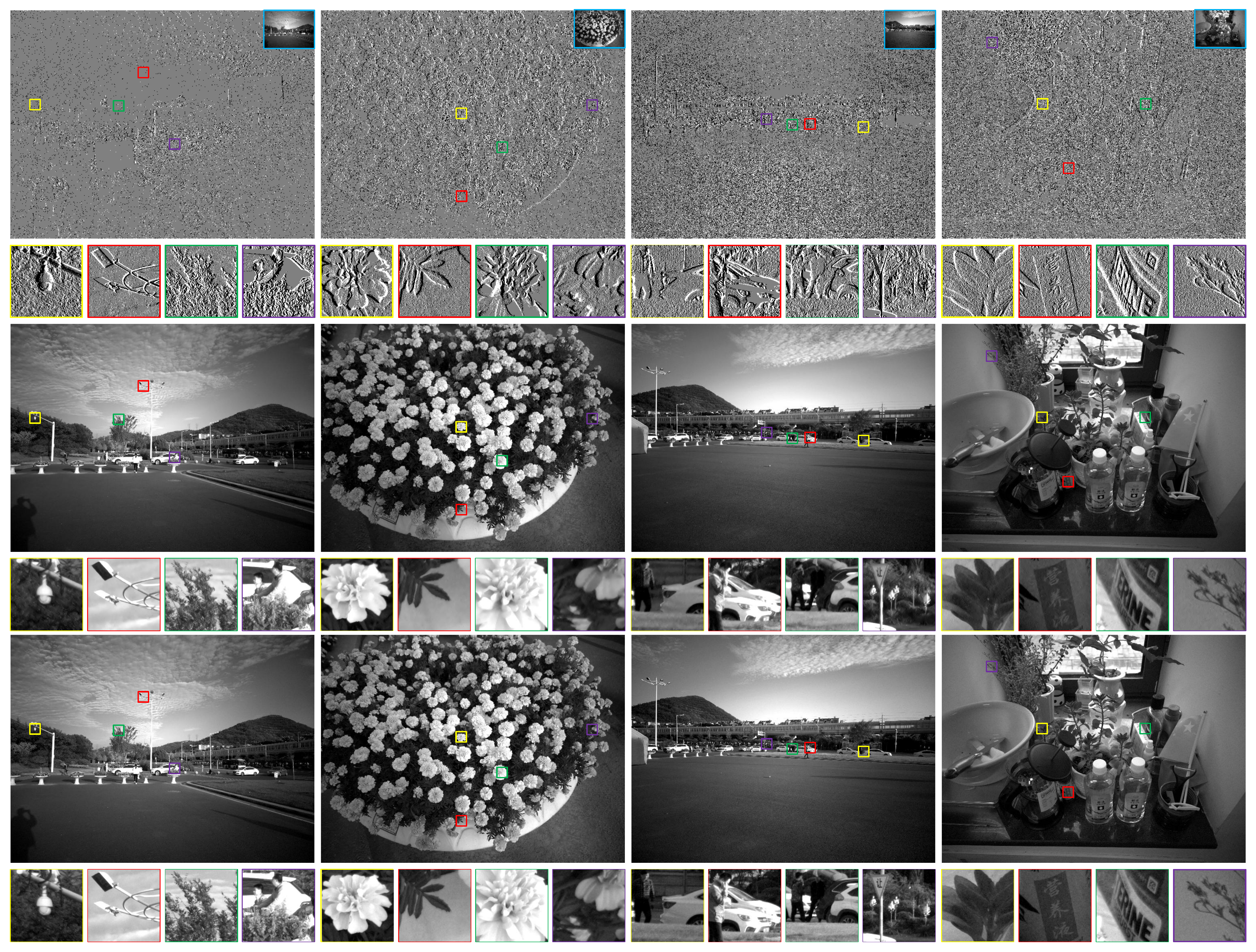}
\caption{Real captured images, compressive ratio: 0.066. HRG map: top, reconstruction result: middle, groundtruth: bottom.}
\label{fig: xiaomi}
\end{figure*}

To demonstrate the effectiveness of the proposed method, we test our method on the High resolution dataset, i.e. the PANDA dataset ~\cite{wang2020panda}, which is a high resolution video dataset, containing $5$ high resolution video sequence in testset. Fig.~\ref{fig: gigapixels} shows the reconstruction results on a 800 megapixel image. Specifically, with the groundtruth image, we first synthesize the HRG map with thresholding $\{-4, 4\}$ and downsample the image with $8\times8$ average pooling to get the LRI. To test the reconstruction fidelity, we divide the gradient image into $2048\times 2048$ patches, and the corresponding LRI image is divided into $256\times 256$ patches. The reconstructed high resolution image are tilled together with the reconstructed patches and the reconstruction fidelity is measured with the groundtruth and recovered image. In average, the compressive ratio of PANDA dataset is 0.0103 and the reconstruction fidelity is $44.37$ dB for psnr and $0.984$ for ssim. In all, experiments on the high resolution dataset demonstrate that high quality imaging with low data transmission bandwidth with the proposed method.

\paragraph{Ablation study}

\begin{figure}[htbp]
\centering
\includegraphics[width=0.5\textwidth]{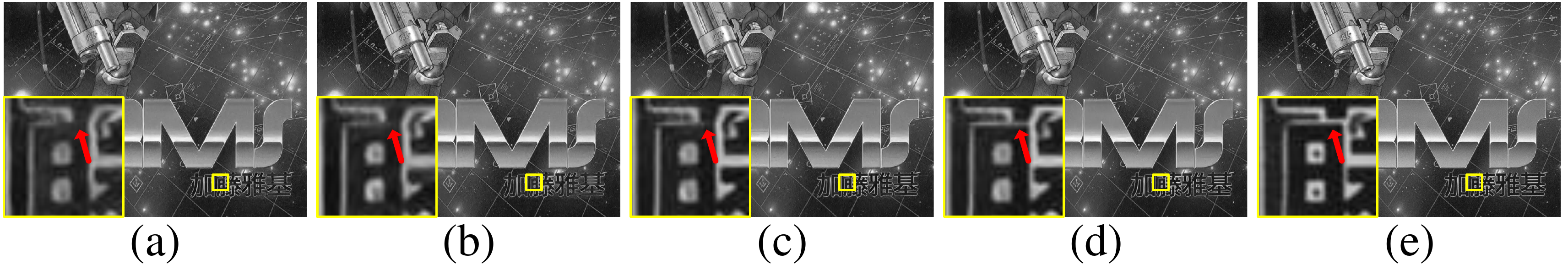}
\caption{Ablation study. (a). without LRI to HRG fusion connection, (b). without multi-scale in gradient branches, (c). without gradient branch in $x$ direction, (d). full model, (e). groundtruth.}
\label{fig: image_ablation_study}
\end{figure}

In order to verify the effectiveness of the proposed neural network, we implement ablation experiments to demonstrate the effectiveness of the information fusion from LRI to HRG, multi-scale structure of the gradient branch and the second gradient branch that reconstructing the gradient in the other direction (y-direction) compared with the direction of the captured gradient image (x-direction). The experimental results are shown in Table~\ref{tab:ablation study} and Fig.~\ref{fig: image_ablation_study}. Through comparison, we could find that the information fusion and multi-scale structure play an important role in improving the reconstruction quality and only when both of them are contained in the network, the effect could be maximized, i.e. around 2 dB improvement in PSNR. Through further comparing the reconstruction in Fig.~\ref{fig: image_ablation_study}(c) and (e), we could find out that without the second gradient reconstruction block, the vertical line cannot be retrieved due to the absence of the gradient information along the y direction. 
With the second gradient branch, the gradient information along the y direction can be predicted from the x-direction gradient image and complement the reconstruction of the vertical details of the scene.

\begin{table}[htbp]
\centering
\caption{Experiental results of ablation study.}
	\scalebox{0.8}{
	\begin{tabular}{c|cccc}
	\hline
	\multicolumn{1}{c|}{\multirow{2}[0]{*}{Methods}} &
				\multicolumn{1}{c}{Without} & \multicolumn{1}{c}{Without} & \multicolumn{1}{c}{Without} & \multicolumn{1}{c}{Full}\\
				& \multicolumn{1}{c}{fusion} & \multicolumn{1}{c}{multi-scale} & \multicolumn{1}{c}{one branch} & \multicolumn{1}{c}{model}\\
	\hline
	manga109 & 29.68/0.936 & 29.50/0.934 & 31.03/0.946 & 31.31/0.948 \\
	\hline
	stanfordLF & 34.59/0.952 & 34.25/0.949 & 35.36/0.958 & 35.79/0.960 \\
	\hline
	YUP++  & 31.13/0.921 & 30.88/0.920 & 31.81/0.931 & 31.95/0.934 \\
	\hline
	\end{tabular}
				}
	\label{tab:ablation study}
\end{table}

\subsection{Experiental results on real captured images}

In order to verify the high throughput imaging on real data, we adopt the camera of Mi10 with Sansung ISOCELL Bright HMX image sensor \cite{samsung}. The size of the sensor is 1/1.33 inches with 0.8 um pixel size and 108 megapixel. With the camera, we could collect the real data and reconstruct the high throughput images with the proposed method. Fig.~\ref{fig: xiaomi} shows our results on real data with the corresponding gradient image, $8\times$ low resolution image (blue insets on the top left of the gradient image), the reconstructed images with the proposed method and the groundtruth image. As shown in the small insets of different region of the captured data, the gradient image could retain most of the details of the scene, such as the network camera for surveillance, the street lamp, the tree, people, \etal, both for indoor scene and outdoor scene. Through combining the high frequency information captured with the gradient image with the low frequency information from the low resolution image, high resolution image could be reconstructed with high fidelity. 
Furthermore, since the gradient image is highly sparse, with the proposed simple compression method, the average compression ratio of the gradient image can reach 0.0606, demonstrating the efficiency and effectiveness of the proposed imaging scheme.

%% file: conclusion.tex
\section{Conclusion}
In this paper, we explore the high throughput imaging scheme based on the combination of low resolution image and low bit gradient image to overcome the readout and transmission bottleneck to realize compact high throughput video acquisition. A modified run-length and entropy coding module is developed and achieved less than 10\% compression ratio, enabling the data transmission with existing output transmission interface. We propose a CNN-based reconstruction model which can effectively reconstruct the low frequency information, structures and details of scenes at high accuracy. The effectiveness and efficiency of the proposed method are demonstrated with both high resolution dataset and real data.